\documentclass[letterpaper]{article} % DO NOT CHANGE THIS
\usepackage{aaai24}  % DO NOT CHANGE THIS
\usepackage{times}  % DO NOT CHANGE THIS
\usepackage{helvet}  % DO NOT CHANGE THIS
\usepackage{courier}  % DO NOT CHANGE THIS
\usepackage[hyphens]{url}  % DO NOT CHANGE THIS
\usepackage{graphicx} % DO NOT CHANGE THIS
\usepackage{natbib}  % DO NOT CHANGE THIS AND DO NOT ADD ANY OPTIONS TO IT
\usepackage{caption} % DO NOT CHANGE THIS AND DO NOT ADD ANY OPTIONS TO IT
\usepackage{algorithm}
\usepackage{algorithmic}
\usepackage{multirow}
\usepackage{subfigure}
\usepackage{amsmath,amssymb}
\usepackage{algorithm}
\usepackage{algorithmic}

\usepackage{booktabs}       % professional-quality 

\usepackage{newfloat}
\usepackage{listings}
\DeclareCaptionStyle{ruled}{labelfont=normalfont,labelsep=colon,strut=off} % DO NOT CHANGE THIS
\floatstyle{ruled}
\newfloat{listing}{tb}{lst}{}
\floatname{listing}{Listing}
\nocopyright
\title{Boosting the Cross-Architecture Generalization of Dataset Distillation through an Empirical Study}
\author{Lirui Zhao$^1$, Yuxin Zhang$^1$, Fei Chao$^{1}$, Rongrong Ji$^1$\thanks{Corresponding Author}}

\affiliations{$^1$Key Laboratory of Multimedia Trusted Perception and Efficient Computing, \\
Ministry of Education of China, School of Informatics, Xiamen University, Xiamen, China \\
{\tt\small liruizhao@stu.xmu.edu.cn, yuxinzhang@stu.xmu.edu.cn}\\ 
{\tt\small fchao@xmu.edu.cn, rrji@xmu.edu.cn}}

\usepackage{bibentry}
\usepackage{pifont}   
\usepackage{subcaption}

\newcommand{\etal}{\textit{et al}.}
\newcommand{\eg}{\textit{e.g.}}
\newcommand{\cmark}{\ding{51}}          % V
\begin{document}

\maketitle

\begin{abstract}
The poor cross-architecture generalization of dataset distillation greatly weakens its practical significance.
This paper attempts to mitigate this issue through an empirical study, which suggests that the synthetic datasets undergo an inductive bias towards the distillation model.
Therefore, the evaluation model is strictly confined to having similar architectures of the distillation model.
We propose a novel method of EvaLuation with distillation Feature (ELF), which utilizes features from intermediate layers of the distillation model for the cross-architecture evaluation.
In this manner, the evaluation model learns from bias-free knowledge therefore its architecture becomes unfettered while retaining performance.
By performing extensive experiments, we successfully prove that ELF can well enhance the cross-architecture generalization of current DD methods.
Code of this project is at \url{https://github.com/Lirui-Zhao/ELF}.
\end{abstract}

\section{Introduction}\label{introduction}
Recent years have witnessed encouraging progress of deep neural networks (DNNs) across a wide range of vision applications, such as image classification~\cite{szegedy2015going,he2016deep}, object detection~\cite{girshick2014rich,ren2015faster}, instance segmentation~\cite{long2015fully,chen2017deeplab} and many others.
Nevertheless, the success of DNNs heavily counts on very large-scale datasets~\cite{deng2009imagenet,lin2014microsoft} that generally contain millions of samples and pose enormous computational burden in consequence.
By mimicking the original large-scale datasets with a smaller size of synthesized samples, dataset distillation (DD)~\cite{wang2018dataset, zhao2020dataset} has therefore been a topic of active research interest lately to maintain model performance as well as to relieve training burden.
The most common metric to evaluate DD methods attributes to the training performance on the generated synthetic datasets.
In contemporary methods~\cite{zhao2021dataset,cazenavette2022dataset}, the networks for distilling dataset and evaluating performance stay the same of a 3-/4-/5-layer shallow ConvNet~\cite{gidaris2018dynamic}.
%

%%%%%%%%%%%%%%%%%%%%%%%%%%%%%%%%%%%%%%%%%%%%%%%%%%%%%%%%%%%
\begin{figure}[!t]
\begin{center}
\includegraphics[width = \linewidth]{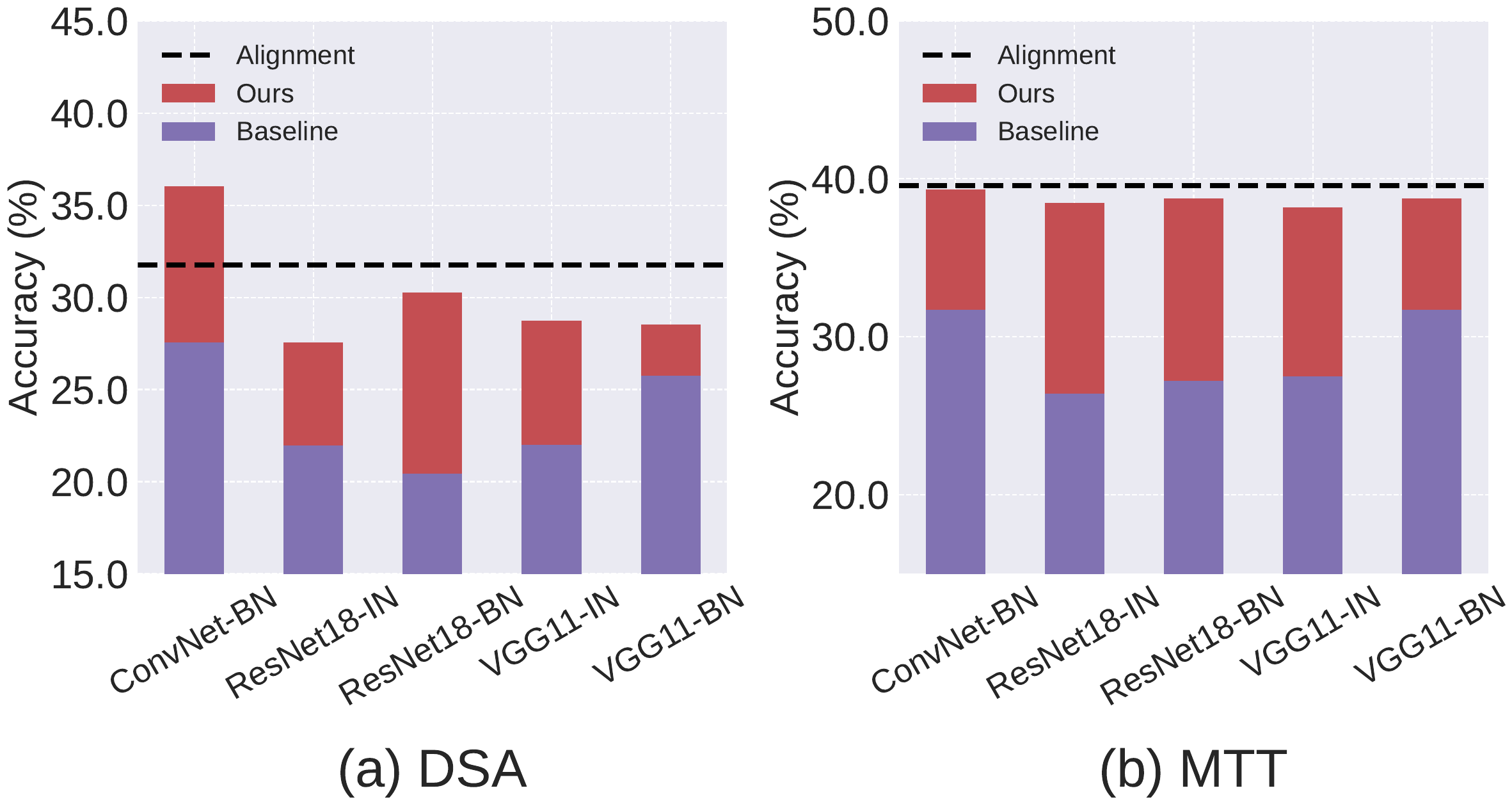}
\end{center}
\vspace{-0.3cm}
\caption{
Improvement of the proposed ELF over existing baseline methods including (a) DSA and (b) MTT. Here, the distillation model is Conv-IN with width of 128 and depth of 3. ``IN'' and ``BN'' denote instance normalization and batch normalization. The horizontal coordinates show the different evaluation models. ``Alignment'' stands for the same evaluation model with the distillation model. Experiments are performed on CIFAR-100 dataset with 10 images per class (IPC).
}
\label{fig:ELF_improvement_cross}
\vspace{-0.3cm}
\end{figure}
%%%%%%%%%%%%%%%%%%%%%%%%%%%%%%%%%%%%%%%%%%%%%%%%%%%%%%%%%%%

%%%%%%%%%%%%%%%%%%%%%%%%%%%%%%%%%%%%%%%%%%%%%%%%%%%%%%%%%%%
\begin{figure*}[!t]
\begin{center}
\includegraphics[width = 0.75\linewidth]{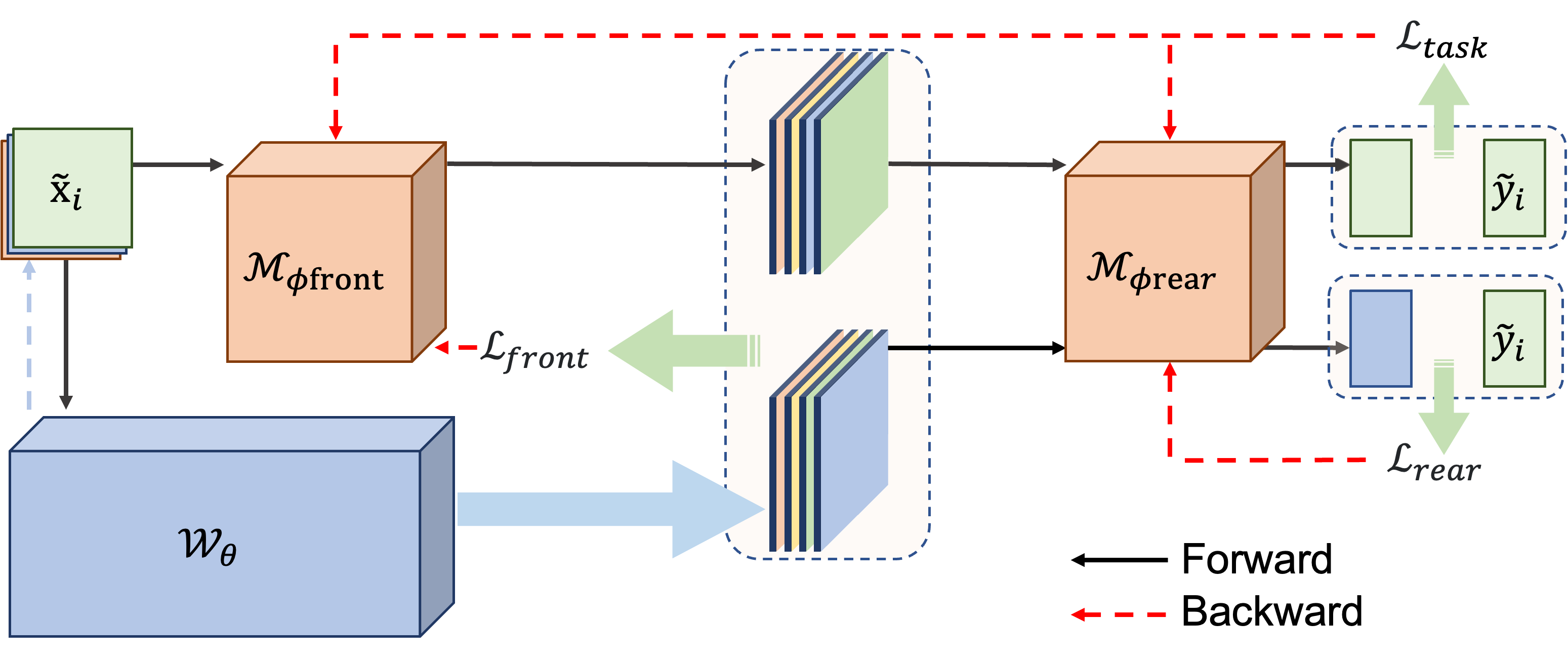}
\end{center}
\vspace{-0.3cm}
\caption{Framework of our proposed ELF method.
We feed the synthetic dataset to the distillation model and obtain the bias-free intermediate features, which are then used to guide the training process of the evaluation model.
}
\label{fig:method}
\vspace{-0.3cm}
\end{figure*}
%%%%%%%%%%%%%%%%%%%%%%%%%%%%%%%%%%%%%%%%%%%%%%%%%%%%%%%%%%%

%
Despite the continuous advances in this setting, the cross-architecture generalization, referring to as individually distilling and evaluating the synthetic dataset with networks of different architectures, is rather disregarded by the community.
We definitely expect a robust performance gain from varying networks trained upon the synthetic dataset while reducing training burden.
Unfortunately, a unified distillation and evaluation network fails to achieve cross-architecture generalization. For example, ResNet-18~\cite{he2016deep}, a strong model that achieves 92.6\% on CIFAR-10 dataset, reportedly in~\cite{cazenavette2022dataset}, has only 46.4\% on the synthetic dataset distilled by ConvNet.
This huge performance gap impedes the practical significance of DD methods.
We therefore in this paper aim at alleviating the poor cross-architecture generalization of existing DD methods with our endeavors primarily involving two empirical studies:
First, networks of more architectural similarity lead to less performance gap between evaluation and distillation models.
Second, performance benefits more from inserting an identical normalization layer with the distillation model into the evaluation model.
The presumable reason lies in that the synthetic dataset retains an inductive bias towards the distillation model, which is instead relieved by similar architectures and identical normalization since they align the distillation and evaluation models.
Therefore, the inductive bias imprisons the cross-architecture generalization.
%

%%%%%%%%%%%%%%%%%%%%%%%%%%%%%%%%%%%%%%%%%%%%%%%%%%%%%%%%%%%
%\begin{figure*}[!t]
%\begin{center}
%\includegraphics[width = 0.9\linewidth]{Figue/result_histogram.pdf}
%\end{center}
%\vspace{-0.3cm}
%\caption{
%Improvement of the proposed ELF over existing baseline methods including (a) DM~\cite{zhao2023dataset}, (b) DSA~\cite{zhao2021dataset} and (c) MTT~\cite{cazenavette2022dataset}. Here, the distillation model is Conv-IN with width of 128 and depth of 3. ``IN'' and ``BN'' denote instance normalization~\cite{ulyanov2016instance} and batch normalization~\cite{ioffe2015batch}. The horizontal coordinates show the different evaluation models. ``Alignment'' stands for the same evaluation model with the distillation model. Experiments are performed on CIFAR-100 dataset~\cite{krizhevsky2009learning} with 10 images per class (IPC).
%}
%\label{fig:ELF_improvement_cross}
%\vspace{-0.3cm}
%\end{figure*}
%%%%%%%%%%%%%%%%%%%%%%%%%%%%%%%%%%%%%%%%%%%%%%%%%%%%%%%%%%%

%
We go over in this paper and realize two innate advantages in the distillation model:
\textit{First}, the architecture of distillation model is by nature popular with the synthesized dataset.
\textit{Second}, the feature maps from distillation model are invulnerable to the inductive bias.
Therefore, we propose a novel method of EvaLuation with distillation Feature (ELF) to boost the cross-architecture generalization of existing DD methods. Fig.\,\ref{fig:method} outlines the framework of our proposed ELF. As its name suggests, we capitalize fully on the bias-free intermediate features of the distillation model as an auxiliary to train the evaluation model from two perspectives:
\textit{First}, the distillation features are leveraged as a form of supervision for the intermediate outputs of evaluation model.
\textit{Second}, the distillation features play as disturbed variables to refine the performance of evaluation model in predicting labels.
Note that, the trained distillation model is widely available from most existing DD methods, which makes our ELF a plug-and-play method to improve the cross-architecture performance.
For example, Fig.\,\ref{fig:ELF_improvement_cross} shows great performance enhancement of the proposed ELF on CIFAR-100~\cite{krizhevsky2009learning} dataset over existing state-of-the-art DD methods including DSA~\cite{zhao2021dataset} and MTT~\cite{cazenavette2022dataset}.
More results on CIFAR-10/100~\cite{krizhevsky2009learning}, Tiny ImageNet~\cite{le2015tiny}, and ImageNet~\cite{russakovsky2015imagenet} are provided in Sec.\,\ref{experiment}, which substantially prove the efficacy of the proposed ELF method.
The main contributions we have made in this paper include:
\textit{First}, an empirical study to reveal the inductive bias in poor cross-architecture generalization. 
\textit{Second}, a novel method of evaluation with distillation feature to boost the cross-architecture generalization.
\textit{Third}, significant performance improvement on the basis of existing methods for dataset distillation.

%%%%%%%%%%%%%%%%%%%%%%%%%%%%%%%%%%%%%%%%%%%%%%%%%%%%%%%%%%%
\begin{table*}[!tbp]
\resizebox{\textwidth}{!}{
\centering
\small
\begin{tabular}{lccccccc}
\toprule
\multirow{2}{*}{} & \multirow{2}{*}[-2pt]{\textbf{Distillation Model}} & \multicolumn{6}{c}{\textbf{Evaluation Model} (\%)} \\ 
\cmidrule{3-8}
 &  & ConvNet-IN & ConvNet-BN & ResNet18-IN & ResNet18-BN & VGG11-IN & VGG11-BN \\ \midrule
\multirow{2}{*}{DM~\cite{zhao2023dataset}} & \textbf{ConvNet-IN} & 49.52 $\pm$ 0.19 & 46.17 $\pm$ 0.52 & 37.38 $\pm$ 2.34 & 39.41 $\pm$ 0.49 & 41.06 $\pm$ 0.71 & 43.80 $\pm$ 0.41 \\
 & ConvNet-BN & 42.13 $\pm$ 0.79 & 49.73 $\pm$ 0.44 & 36.71 $\pm$ 1.42 & 42.51 $\pm$ 0.81 & 36.11 $\pm$ 0.51 & 44.74 $\pm$ 0.30 \\ \midrule
\multirow{2}{*}{DSA~\cite{zhao2021dataset}} &\textbf{ConvNet-IN} & 51.70 $\pm$ 0.36 & 43.25 $\pm$ 0.71 & 41.98 $\pm$ 0.85 & 37.98 $\pm$ 0.88 & 42.98 $\pm$ 0.81 & 42.66 $\pm$ 0.67 \\
 & ConvNet-BN & 34.79 $\pm$ 0.34 & 45.84 $\pm$ 0.69 & 31.43 $\pm$ 0.73 & 33.13 $\pm$ 0.68 & 29.81 $\pm$ 0.32 & 36.20 $\pm$ 0.35 \\ \midrule
\multirow{2}{*}{MTT~\cite{cazenavette2022dataset}} & \textbf{ConvNet-IN} & 63.48 $\pm$ 0.58 & 47.27 $\pm$ 1.20 & 44.72 $\pm$ 1.43 & 42.32 $\pm$ 0.40 & 49.04 $\pm$ 0.50 & 46.95 $\pm$ 1.27 \\
 & ConvNet-BN & 50.50 $\pm$ 0.80 & 54.18 $\pm$ 1.13 & 39.77 $\pm$ 0.71 & 40.94 $\pm$ 2.88 & 44.96 $\pm$ 0.95 & 48.32 $\pm$ 1.82 \\ \midrule
FrePo~\cite{zhou2022dataset} & \textbf{ConvNet-BN} & - & 65.6 $\pm$ 0.6 & 47.4 $\pm$ 0.7 & 53.0 $\pm$ 1.0 & 35.0 $\pm$ 0.7 & 56.8 $\pm$ 0.6 \\ \midrule
Entire CIFAR-10 &  & 86.64 & 88.49 & 92.60 & 93.69 & 88.05 & 90.46 \\ \bottomrule
\end{tabular}}

\caption{Cross-architecture performance on CIFAR-10~\cite{krizhevsky2009learning}.  ConvNet is of width:128 and depth:3. We synthesize 10 images per class (IPC) and the experiments are run over 5 times.}
\label{table:cross_cifar10_10ipc}

\end{table*}
%%%%%%%%%%%%%%%%%%%%%%%%%%%%%%%%%%%%%%%%%%%%%%%%%%%%%%%%%%%

%%%%%%%%%%%%%%%%%%%%%%%%%%%%%%%%%%%%%%%%%%%%%%%%%%%%%%%%%%%
\begin{table*}[]
\small
\centering
\begin{tabular}{lcccc}
\toprule
\multicolumn{1}{c}{} & ConvNet-IN & ConvNetW256-IN & ConvNetD4-IN & ResNet18-IN \\ \midrule
Synthetic (\%) & 63.48 $\pm$ 0.58 & 64.75 $\pm$ 0.20 & 59.03 $\pm$ 0.65 & 44.72 $\pm$ 1.43 \\
CIFAR-10 (\%) & 86.64 & 88.27 & 87.20 & 92.36 \\ \midrule
Synthetic (\%) & 38.16 $\pm$ 0.32 & 39.91 $\pm$ 0.32 & 34.11 $\pm$ 0.27 & 26.33 $\pm$ 0.52 \\
CIFAR-100 (\%) & 57.78 & 61.99 & 59.24 & 66.50 \\ \bottomrule
\end{tabular}
\caption{
Performance of evaluation models with similar architectures to the distillation model ConvNet-IN~\cite{gidaris2018dynamic}. We utilize MTT~\cite{cazenavette2022dataset} to synthesize a dataset (10 IPC) from CIFAR-10/100~\cite{krizhevsky2009learning}
"W256" and "D4" denote the width of 256 channels and the depth of 4 layers, respectively.
}. 
\label{table:cross_MTT_similarity}
\vspace{-0.5cm}
\end{table*}
%%%%%%%%%%%%%%%%%%%%%%%%%%%%%%%%%%%%%%%%%%%%%%%%%%%%%%%%%%%

\section{Related work}
As a pioneer in dataset distillation (DD), Wang \emph{et al}.~\cite{wang2018dataset} proposed to synthesize a few number of image samples to achieve comparable training performance as an alternative to the original large-scale dataset. Following this footprint, abundant consecutive studies that can be broadly classified into three groups, have been introduced~\cite{sachdeva2023data,lei2023comprehensive,yu2023dataset}.

The first group encourages every single- or multiple-step parameter consistency between models trained upon the synthetic and original datasets.
To this end, Zhao \emph{et al}.~\cite{zhao2020dataset} proposed to match per-step gradients between distillation models from the synthetic and original datasets. Some followers~\cite{zhao2021dataset,lee2022dataset} take a further step by including differentiable Siamese augmentation or contrastive signals.
Instead, Cazenavette~\etal~\cite{cazenavette2022dataset} revealed error accumulation from the single-step gradient matching, and proposed matching training trajectories (MTT) to align parameters in every multiple steps. The discrepancy between the distillation and subsequent may cause accumulated trajectory error, therefore, Du~\etal~\cite{du2022minimizing} further proposed to seek a ﬂat trajectory. 
Cui~\etal~\cite{cui2022scaling} re-parameterized parameter-matching loss in MTT to reduce the memory requirement.
They further proposed to assign soft labels for synthetic datasets when it comes to distilling datasets with a larger number of categories (\eg, ImageNet-1K~\cite{russakovsky2015imagenet}).
The second group uplifts the dataset alignment in the distribution level. For example, Wang~\etal~\cite{wang2022cafe} aligned layer-wise features between the real and synthetic datasets, and encoded the discriminant power into the synthetic clusters.
Zhao~\etal~\cite{zhao2023dataset} minimized the difference in the embedding space.
The last group takes into consideration the Meta-learning~\cite{finn2017model} for guiding DD in a bi-level optimization fashion. 
In~\cite{nguyen2020dataset,nguyen2021dataset}, the inner optimization is approximated using neural tangent kernel~\cite{lee2019wide}.
Zhou~\etal~\cite{zhou2022dataset} proposed neural feature regression with pooling to alleviate the heavy computation and memory costs from inner loop learning of Meta-learning.
The back-propagation through time method~\cite{werbos1990backpropagation} is used to recursively compute the meta-gradient and update the synthetic dataset~\cite{wang2018dataset}. It is also considered to optimize image labels~\cite{bohdal2020flexible,sucholutsky2021soft}.
Deng~\etal~\cite{deng2022remember} showed a momentum term and a longer unrolled trajectory in the inner loop well enhance distillation performance.

These DD methods mostly focus on evaluating performance of synthetic datasets within the same architecture. This paper prioritizes enhancing the cross-architecture performance. Besides, recent works~\cite{liu2022dataset,deng2022remember,kim2022dataset,lee2022latent} factorize synthetic datasets to increase samples of more diversity and less redundancy.
Our effort in this paper is directly towards analyzing and improving the cross-architecture performance of the non-factorized methods, leading to a complementary method to increasing synthetic samples.
%-------------------------------------------------------------------------

%%%%%%%%%%%%%%%%%%%%%%%%%%%%%%%%%%%%%%%%%%%%%%%%%%%%%%%%%%%
\begin{table*}[t]
\centering
\resizebox{\textwidth}{!}{
\begin{tabular}{lccccccc}
\toprule
\multirow{2}{*}{} & \multirow{2}{*}[-2pt]{\textbf{Distillation Model}} & \multicolumn{6}{c}{\textbf{Evaluation Model} (\%)} \\ \cmidrule{3-8} 
 &  & ConvNet-IN & ConvNet-BN & ResNet18-IN & ResNet18-BN & VGG11-IN & VGG11-BN \\ \midrule
\multirow{2}{*}{DSA~\cite{zhao2021dataset}} & \textbf{ConvNet-IN} & 61.14 $\pm$ 0.30 & 56.89 $\pm$ 0.21 & 49.50 $\pm$ 0.49 & 50.71 $\pm$ 0.54 & 51.11 $\pm$ 0.16 & 55.80 $\pm$ 0.44 \\
  & ConvNet-BN & 48.49 $\pm$ 0.62 & 60.44 $\pm$ 0.47 & 42.61 $\pm$ 0.20 & 47.87 $\pm$ 1.30& 43.11 $\pm$ 0.43 & 51.63 $\pm$ 0.38 \\ \midrule
\multirow{2}{*}{MTT~\cite{cazenavette2022dataset}} & \textbf{ConvNet-IN} & 71.60 $\pm$ 0.20 & 62.65 $\pm$ 0.60 & 57.68 $\pm$ 0.71 & 58.48 $\pm$ 0.89 & 62.09 $\pm$ 0.40 & 63.31 $\pm$ 0.50 \\
  & ConvNet-BN & 62.43 $\pm$ 0.27 & 69.50 $\pm$ 0.89 & 53.79 $\pm$ 0.85 & 61.84 $\pm$ 0.89 & 58.01 $\pm$ 1.00 & 64.69 $\pm$ 0.68 \\ \midrule
Entire CIFAR-10 &  & 86.64 & 88.49 & 92.39 & 93.69 & 88.05 & 90.46 \\ \bottomrule
\end{tabular}
}
\caption{Cross-architecture performance on CIFAR10 with 50 IPC. For DSA~\cite{zhao2021dataset}, ResNet18~\cite{he2016deep} and VGG11~\cite{szegedy2015going} are in favor of batch normalization.}
\label{table:cross_cifar10_50ipc}
\end{table*}
%%%%%%%%%%%%%%%%%%%%%%%%%%%%%%%%%%%%%%%%%%%%%%%%%%%%%%%%%%%

%%%%%%%%%%%%%%%%%%%%%%%%%%%%%%%%%%%%%%%%%%%%%%%%%%%%%%%%%%
\begin{table*}[t]
\centering
\resizebox{\textwidth}{!}{
\begin{tabular}{lccccccc}
\toprule
\multirow{2}{*}{MTT~\cite{cazenavette2022dataset}} & \multirow{2}{*}[-2pt]{\textbf{Distillation Model}} & \multicolumn{6}{c}{\textbf{Evaluation Model} (\%)} \\ \cmidrule{3-8} 
 &  & ConvNet-IN & ConvNet-BN & ResNet18-IN & ResNet18-BN & VGG11-IN & VGG11-BN \\ \midrule
\multirow{2}{*}{10 IPC} & \textbf{ConvNet-IN} & 39.58 $\pm$ 0.24 & 31.73 $\pm$ 0.15 & 26.39 $\pm$ 0.66 & 27.21 $\pm$ 0.53 & 27.50 $\pm$ 0.26 & 31.71 $\pm$ 0.58 \\
  & ConvNet-BN & 30.16 $\pm$ 0.32 & 36.78 $\pm$ 0.18 &21.46 $\pm$ 0.62 & 27.24 $\pm$ 0.69 & 23.10 $\pm$ 0.28 & 31.35 $\pm$ 0.69 \\ \midrule
\multirow{2}{*}{50 IPC} & \textbf{ConvNet-IN} & 47.03 $\pm$ 0.15 & 47.27 $\pm$ 0.19 & 41.17 $\pm$ 0.52 & 46.43 $\pm$ 0.45 & 41.59 $\pm$ 0.19 & 49.02 $\pm$ 0.19 \\
 & ConvNet-BN & 44.76 $\pm$ 0.17 & 51.11 $\pm$ 0.32 & 40.59 $\pm$ 0.35 & 49.49 $\pm$ 0.48 & 38.97 $\pm$ 0.33 & 50.72 $\pm$ 0.25 \\ \midrule
Entire CIFAR-100 &  & 57.78 & 63.09 & 66.50 & 74.75 & 56.72 & 68.06 \\ \bottomrule
\end{tabular}}
\caption{Cross-architecture performance of MTT~\cite{cazenavette2022dataset} on CIFAR100~\cite{krizhevsky2009learning}. A large dataset benefits more to models with batch normalization.}
\label{table:cross_cifar100_10/50ipc}
\vspace{-10pt}
\end{table*}
%%%%%%%%%%%%%%%%%%%%%%%%%%%%%%%%%%%%%%%%%%%%%%%%%%%%%%%%%%

\section{Methodology}

\subsection{Background}\label{sec:3-1}
Consider a large-scale training dataset $\mathcal{T}=\left\{\left(x_i, y_i\right)\right\}_{i=1}^{|\mathcal{T}|}$ that contains $|\mathcal{T}|$ pairs of training sample $x_i$ and corresponding label $y_i$, as well as a testing dataset $\mathcal{T}_{test}$ of the same domain with $\mathcal{T}$.
DD excavates a distillation model $\mathcal{W}_{\boldsymbol{\theta}}$ to condense $\mathcal{T}$ into a small-scale version $\mathcal{S}=\left\{\left(\tilde{x}_i, \tilde{y}_i\right)\right\}_{i=1}^{|\mathcal{S}|}$ where $|\mathcal{S}|<<|\mathcal{T}|$, such that, when testing on $\mathcal{T}_{test}$, an evaluation model $\mathcal{M}_{\boldsymbol{\phi}}$ efficiently trained on $\mathcal{S}$ performs on par with that cumbersomely trained on $\mathcal{T}$.
Here, the $\boldsymbol{\theta}$ and $\boldsymbol{\phi}$ are the parameters of $\mathcal{W}$ and $\mathcal{M}$.
Off-the-shelf methods implement this target from three perspectives as we briefly describe in the following.

Parameter-matching based methods separately train the distillation model using $\mathcal{T}$ and $\mathcal{S}$, resulting in $\boldsymbol{\theta}^\mathcal{T}$ and $\boldsymbol{\theta}^\mathcal{S}$. During training, $\mathcal{S}$ is iteratively optimized to match the updating parameters of $\boldsymbol{\theta}^\mathcal{T}$ and $\boldsymbol{\theta}^\mathcal{S}$ as:
\begin{equation}\label{eq:param_matching}
\mathcal{L}(\mathcal{S}) =D(\Delta \boldsymbol{\theta}^\mathcal{T},\Delta \boldsymbol{\theta}^\mathcal{S}),
\end{equation}
where $D(\cdot,\cdot)$ is a specific distance function like cosine similarity, and $\Delta \boldsymbol{\theta}^\mathcal{T}$ stands as gradient accumulation from single~\cite{zhao2021dataset} or multiple~\cite{cazenavette2022dataset} training steps.

Distribution-matching based methods update $\mathcal{S}$ by matching the distributions of $\mathcal{T}$ and $\mathcal{S}$. The distribution is measured with the features from the distillation model~\cite{zhao2023dataset}:
\begin{equation}\label{eq:distribution}
\mathcal{L}(\mathcal{S})= D \Big( \mathcal{F}\big(\mathcal{W}_{\boldsymbol{\theta}}(\mathcal{T})\big), \mathcal{F}\big(\mathcal{W}_{\boldsymbol{\theta}}(\mathcal{S})\big) \Big),
\end{equation}
where $\mathcal{F}(\cdot, \cdot)$ denotes the feature maps from the last convolutional block.

Meta-learning based methods optimize $\mathcal{S}$ in a bi-level manner and can be formulated as:
\begin{equation}\label{eq:meta_learning}
\mathcal{S}^*=\arg \min _{\mathcal{S}} \mathbb{E}_{(x,y) \sim \mathcal{T}}\Big[\mathcal{L}\big(\mathcal{W}_{\boldsymbol{\theta^*}}(x), y\big)\Big], 
\end{equation}
subject to
\begin{equation}
\boldsymbol{\theta}^*=\arg \min _{\boldsymbol{\theta}} \mathbb{E}_{(\tilde{x}, \tilde{y}) \sim \mathcal{S}}\Big[\mathcal{L}\big(\mathcal{W}_{\boldsymbol{\theta}}(\tilde{x}), \tilde{y}\big)\Big],
\end{equation}
where $\mathcal{L}(\cdot, \cdot)$ is the training loss with respect to specific sample pairs.
In this way, $\mathcal{S}$ is updated such that the model trained on $\mathcal{S}$ minimizes the training loss over $\mathcal{T}$.
Currently, most DD methods consider a 3-/4-/5-layer ConvNet as the network backbone for both the distillation and evaluation models.
The community rarely investigates the cross-architecture generalization of DD methods, which uses different networks to respectively distill and evaluate the synthetic dataset. It is natural to expect advanced networks like ResNet~\cite{he2016deep} to perform well on the synthetic dataset distilled by the simple ConvNet.
%-------------------------------------------------------------------------

\subsection{An Empirical Study of Cross-Architecture Generalization}\label{sec:3-2}

We begin by investigating the cross-architecture generalization of existing DD methods through an empirical study.
To derive the small-scale dataset $\mathcal{S}$, we consider representative methods including parameter-matching based DSA~\cite{zhao2021dataset} and MTT~\cite{cazenavette2022dataset}, distribution-matching based DM~\cite{zhao2023dataset}, and meta-learning based FrePo~\cite{zhou2022dataset}.
Table\,\ref{table:cross_cifar10_10ipc} reports the performance for distilling CIFAR-10 dataset~\cite{krizhevsky2009learning} where the distilled small dataset $\mathcal{S}$ consists of 10 images per class (IPC) and the evaluation models are equipped with different architectures over the distillation models.
We have two observations.

\textit{First}, all DD methods suffer notable performance degradation in the cross-architecture evaluation. For example, ResNet18~\cite{he2016deep} with batch normalization layers is reportedly to have 93.69\% top-1 accuracy on CIFAR10, however, only 53.0\% is obtained if trained on $\mathcal{S}$ distilled by ConvNet-BN, as shown in the FrePo method of Table.\,\ref{table:cross_cifar10_10ipc}. These strong modern networks perform even worse than the simple ConvNet.
We attribute the poor performance to the inductive bias of $\mathcal{S}$ towards the distillation model.
As delineated by all Eq.\,(\ref{eq:param_matching}) to Eq.\,(\ref{eq:meta_learning}), the loss function utilized in current DD methods serves as an indirect constraint related to the dynamic training process's model parameters or output distribution.
Though this ensures a robust performance if the same network architecture is adopted for evaluation model, it drives $\mathcal{S}$ towards overfitting the training procedure of the distillation model. Therefore, this phenomenon contributes to the observed deterioration in evaluation performance of $\mathcal{S}$ as the disparity in model architectures widens.

\textit{Second}, the performance drop can be somewhat mitigated through an alignment of normalization layers. For instance, ResNet18-IN shows better accuracy than ResNet18-BN when trained on $\mathcal{S}$ synthesized by ConvNet-IN, and vice versa.
This empirically proves the inductive bias as the same normalization layer indicates a more similar structure between the distillation and evaluation models.
Table\,\ref{table:cross_MTT_similarity} further investigates how the similarity of architectures influences the cross-architecture generalization by evaluating multiple variants of ConvNet on $\mathcal{S}$ distilled by its vanilla version.
We can safely conclude that the performance degradation of the cross-architecture evaluation is closely related to the architecture difference.
The alignment of architecture between the evaluation and distillation models mitigates the impact of inductive bias.
Unfortunately, the requirement of similar architectures confines the cross-architecture generalization. 

Moreover, normalization alignment also incarcerates the capability of modern networks.
The cross-architecture experiment on CIFAR10 with 50 IPC is conducted in Table\,\ref{table:cross_cifar10_50ipc}.
Light anomalies, such as those in DSA where ResNet18 and VGG11 favor batch normalization even on the synthetic dataset distilled by ConvNet-IN, can be observed.
We posit that this phenomenon arises from the interplay between variations in modern evaluation models (\eg, ResNet's greater advantage with batch normalization) and the inductive bias, which becomes more prominent as synthetic datasets grow in size.
Table\,\ref{table:cross_cifar100_10/50ipc} presents additional results that further validate our analysis and extend the observed trend.
The persistent presence of the inductive bias continues to yield adverse effects, despite the compensatory or even surpassing impact of model disparities.

Therefore, how to mitigate the inductive bias in existing DD methods without a compromise on architecture adaption remains to be well addressed.

%%%%%%%%%%%%%%%%%%%%%%%%%%%%%%%%%%%%%%%%%%%%%%%%%%%%%%%%%%%
\begin{table*}[]
\centering
\resizebox{\textwidth}{!}{
\begin{tabular}{cccccccc}
\toprule
 & \multicolumn{1}{c}{ConvNet-IN} & \multicolumn{1}{c}{} & ConvNet-BN & ResNet18-IN & ResNet18-BN & VGG11-IN & VGG11-BN \\ \midrule
\multirow{3}{*}{DM~\cite{zhao2023dataset}} & \multicolumn{1}{c}{\multirow{3}{*}{49.52 $\pm$ 0.19}} & \multicolumn{1}{c}{Baseline} & 46.17 $\pm$ 0.52 & 37.38 $\pm$ 2.34 & 39.41 $\pm$ 0.49 & 41.06 $\pm$ 0.71 & 43.80 $\pm$ 0.41 \\ 
 & \multicolumn{1}{c}{} & \multicolumn{1}{c}{w. ELF} & 55.19 $\pm$ 0.44 & 38.81 $\pm$ 0.35 & 41.59 $\pm$ 0.68 & 47.52 $\pm$ 0.56 & 46.43 $\pm$ 1.70 \\ 
 & \multicolumn{1}{c}{} & \multicolumn{1}{c}{Gain} & $+$ 9.02 & $+$ 1.43 & $+$ 2.18 & $+$ 6.46 & $+$ 3.35 \\ \midrule
\multirow{3}{*}{DSA~\cite{zhao2021dataset}} & \multicolumn{1}{c}{\multirow{3}{*}{51.70 $\pm$ 0.36}} & \multicolumn{1}{c}{Baseline} & 43.25 $\pm$ 0.71 & 41.98 $\pm$ 0.85 & 37.98 $\pm$ 0.88 & 42.98 $\pm$ 0.81 & 42.66 $\pm$ 0.67 \\ 
 & \multicolumn{1}{c}{} & \multicolumn{1}{c}{w. ELF} & 54.01 $\pm$ 0.48 & 42.29 $\pm$ 0.40 & 40.45 $\pm$ 1.80 & 49.19 $\pm$ 0.21 & 44.62 $\pm$ 2.15 \\ 
 & \multicolumn{1}{c}{} & \multicolumn{1}{c}{Gain} & $+$ 10.76 & $+$ 1.31 & $+$ 2.47 & $+$ 6.21 & $+$ 1.96 \\ \midrule
\multirow{3}{*}{MTT~\cite{cazenavette2022dataset}} & \multicolumn{1}{c}{\multirow{3}{*}{63.48 $\pm$ 0.58}} & \multicolumn{1}{c}{Baseline} & 47.27 $\pm$ 1.20 & 44.72 $\pm$ 1.43 & 42.32 $\pm$ 0.40 & 49.04 $\pm$ 0.50 & 46.95 $\pm$ 1.27 \\
 & \multicolumn{1}{c}{} & \multicolumn{1}{c}{w. ELF} & 58.42 $\pm$ 1.44 & 55.11 $\pm$ 0.89 & 50.16 $\pm$ 1.11 & 61.23 $\pm$ 0.69 & 55.49 $\pm$ 2.32 \\ 
 & \multicolumn{1}{c}{} & \multicolumn{1}{c}{Gain} & $+$ 11.15 & $+$ 10.39 & $+$ 7.84 & $+$ 12.19 & $+$ 8.54 \\ \midrule
\begin{tabular}[c]{@{}c@{}}Entire CIFAR-10\end{tabular} & 86.64 &  & 88.49 & 92.39 & 93.69 & 88.05 & 90.46 \\ \bottomrule
\end{tabular}}

\caption{Cross-architecture performance comparison of different baseline methods and corresponding ELF built upon. Experiments are performed on CIFAR10 with 10 IPC.}
\label{table:ELF_cross_0}
\end{table*}
%%%%%%%%%%%%%%%%%%%%%%%%%%%%%%%%%%%%%%%%%%%%%%%%%%%%%%%%%%%

%%%%%%%%%%%%%%%%%%%%%%%%%%%%%%%%%%%%%%%%%%%%%%%%%%%%%%%%%%%
\begin{table*}[]
\centering
\resizebox{\textwidth}{!}{
\begin{tabular}{cccccccc}
\toprule
 & \multicolumn{1}{c}{ConvNet-IN} & \multicolumn{1}{c}{} & ConvNet-BN & ResNet18-IN & ResNet18-BN & VGG11-IN & VGG11-BN \\ \midrule
\multirow{3}{*}{DM~\cite{zhao2023dataset}} & \multicolumn{1}{c}{\multirow{3}{*}{29.45 $\pm$ 0.27}} & \multicolumn{1}{c}{Baseline} & 28.46 $\pm$ 0.32 & 20.06 $\pm$ 1.96 & 20.98 $\pm$ 0.68 & 21.42 $\pm$ 0.35 & 26.51 $\pm$ 0.37 \\ 
 & \multicolumn{1}{c}{} & \multicolumn{1}{c}{w. ELF} & 35.74 $\pm$ 0.28 & 25.99 $\pm$ 0.27 & 28.12 $\pm$ 0.86 & 28.90 $\pm$ 0.26 & 29.94 $\pm$ 0.48 \\ 
 & \multicolumn{1}{c}{} & \multicolumn{1}{c}{Gain} & $+$ 7.28 & $+$ 5.93 & $+$ 7.14 & $+$ 7.48 & $+$ 3.43 \\ \midrule
\multirow{3}{*}{DSA~\cite{zhao2021dataset}} & \multicolumn{1}{c}{\multirow{3}{*}{31.76 $\pm$ 0.37}} & \multicolumn{1}{c}{Baseline} & 27.56 $\pm$ 0.18 & 21.96 $\pm$ 0.51 & 20.45 $\pm$ 0.53 & 22.00 $\pm$ 0.34 & 25.73 $\pm$ 0.41 \\ 
 & \multicolumn{1}{c}{} & \multicolumn{1}{c}{w. ELF} & 36.02 $\pm$ 0.35 & 27.54 $\pm$ 0.19 & 30.26 $\pm$ 0.65 & 28.74 $\pm$ 0.11 & 28.54 $\pm$ 1.23 \\ 
 & \multicolumn{1}{c}{} & \multicolumn{1}{c}{Gain} & $+$ 8.46 & $+$ 5.58 & $+$ 9.81 & $+$ 6.74 & $+$ 2.81 \\ \midrule
\multirow{3}{*}{MTT~\cite{cazenavette2022dataset}} & \multicolumn{1}{c}{\multirow{3}{*}{39.58 $\pm$ 0.24}} & \multicolumn{1}{c}{Baseline} & 31.73 $\pm$ 0.15 & 26.39 $\pm$ 0.66 & 27.21 $\pm$ 0.53 & 27.50 $\pm$ 0.26 & 31.71 $\pm$ 0.58 \\
 & \multicolumn{1}{c}{} & \multicolumn{1}{c}{w. ELF} & 39.32 $\pm$ 0.24 & 38.48 $\pm$ 0.14 & 38.76 $\pm$ 0.80 & 38.20 $\pm$ 0.49 & 38.78 $\pm$ 0.84 \\ 
 & \multicolumn{1}{c}{} & \multicolumn{1}{c}{Gain} & $+$ 7.59 & $+$ 12.09 & $+$ 11.55 & $+$ 10.70 & $+$ 7.07 \\ \midrule
\begin{tabular}[c]{@{}c@{}}Entire CIFAR-100\end{tabular} & 57.68 &  & 63.09 & 66.50 & 74.75 & 56.72 & 68.06 \\ \bottomrule
\end{tabular}}

\caption{Cross-architecture performance comparison of different baseline methods and corresponding ELF built upon. Experiments are performed on CIFAR100 with 10 IPC.}
\label{table:ELF_cross_1}
\vspace{-10pt}
\end{table*}
%%%%%%%%%%%%%%%%%%%%%%%%%%%%%%%%%%%%%%%%%%%%%%%%%%%%%%%%%%%

\subsection{Evaluation with the Distillation Feature}\label{sec:3-3}

It is obvious that the inductive bias towards the distillation model only occurs to the synthesized $\mathcal{S}$, instead of the original dataset $\mathcal{T}$ or $\mathcal{T}_{test}$. Not surprisingly, an evaluation model beyond the architecture scope of distillation model, performs poorly if tested on $\mathcal{T}_{test}$ while trained on $\mathcal{S}$. 

Looking back on the distillation model $\mathcal{W}_{\boldsymbol{\theta}}$, we find it by nature is a good assistance, for two advantages, to help a distillation-architecture-beyond evaluation model break away the inductive bias:
\textit{First}, the inductive bias does not harm its performance if tested on $\mathcal{T}_{test}$ while trained on $\mathcal{T}$ since its architecture is popular with the synthesized $\mathcal{S}$. 
\textit{Second}, its feature map outputs are not affected by the inductive bias since it is trained upon the original training dataset $\mathcal{T}$.

Motivated by this, we propose a novel EvaLuation-with-the-distillation-Feature method (ELF) to reduce the inductive bias with no sacrifice of various architectures of the evaluation model.
To be concrete, for $\mathcal{S}=\{(\tilde{x}_i, \tilde{y}_i)\}_{i=1}^{|\mathcal{S}|}$, we feed each synthesized image $\tilde{x}_i$ to the distillation model, and obtain output from the last convolutional block as the desired intermediate feature, denoted as $\mathcal{F}\big(\mathcal{W}_{\boldsymbol{\theta}}(\tilde{x}_i)\big)$, which as discussed before, is not influenced by the inductive bias. Therefore, we can utilize it to guide the training process of the evaluation model to mitigate the performance drop.

For ease of the following representation, we split the evaluation model $\mathcal{M}_{\boldsymbol{\phi}}$ into the front section $\mathcal{M}_{\boldsymbol{{\phi}_{front}}}$ and rear section $\mathcal{M}_{\boldsymbol{{\phi}_{rear}}}$, where $\mathcal{M}_{\boldsymbol{{\phi}_{front}}} \cap \mathcal{M}_{\boldsymbol{{\phi}_{rear}}} = \varnothing$ and $\mathcal{M}_{\boldsymbol{{\phi}_{front}}} \cup \mathcal{M}_{\boldsymbol{{\phi}_{rear}}} = \mathcal{M}_{\boldsymbol{{\phi}}}$.
Given an input image $x$, we have:
\begin{equation}
    \mathcal{M}_{\boldsymbol{{\phi}}}(x) = \mathcal{M}_{\boldsymbol{{\phi}_{rear}}}\big(\mathcal{M}_{\boldsymbol{{\phi}_{front}}}(x)\big) .
\end{equation}

We make full use of the bias-free intermediate feature $\mathcal{F}\big(\mathcal{W}_{\boldsymbol{\theta}}(\tilde{x}_i)\big)$ to supervise the output of the front section $\mathcal{M}_{\boldsymbol{{\phi}_{front}}}$ as:
\begin{equation}\label{eq:front_loss}
    \mathcal{L}_{front} = \sum_{i=1}^{|\mathcal{S}|} D\Big( \mathcal{F}\big(\mathcal{W}_{\boldsymbol{\theta}}(\tilde{x}_i)\big), \mathcal{M}_{\boldsymbol{{\phi}_{front}}}(\tilde{x}_i)\Big).
\end{equation}

Also, we feed $\mathcal{F}\big(\mathcal{W}_{\boldsymbol{\theta}}(\tilde{x}_i)\big)$ to the rear section $\mathcal{M}_{\boldsymbol{{\phi}_{rear}}}$ for classification supervision as:
\begin{equation}\label{eq:rear_loss}
    \mathcal{L}_{rear} = \sum_{i=1}^{|\mathcal{S}|} KL \bigg( \mathcal{M}_{\boldsymbol{{\phi}_{rear}}}\Big(\mathcal{F}\big(\mathcal{W}_{\boldsymbol{\theta}}(\tilde{x}_i)\big)\Big) ,\tilde{y}_i\bigg),
\end{equation}
where $KL(\cdot, \cdot)$ returns the cross-entropy loss. This allows the evaluation model to learn to predict labels on the basis of disturbed features.
In addition, the evaluation model itself also learns classification task as:
\begin{equation}\label{eq:task_loss}
    \mathcal{L}_{task} = \sum_{i=1}^{|\mathcal{S}|} KL\big(\mathcal{M}_{\boldsymbol{{\phi}}}(x_i), \tilde{y}_i\big).
\end{equation}

Finally, the training objective of our proposed ELF becomes:
\begin{equation}\label{eq:loss}
    \mathcal{L} = \mathcal{L}_{task} + \lambda_{front}\mathcal{L}_{front} + \lambda_{rear}\mathcal{L}_{rear},
    \end{equation}
in which $\lambda_{front}$ and $\lambda_{rear}$ are two tradeoff parameters.

%%%%%%%%%%%%%%%%%%%%%%%%%%%%%%%%%%%%%%%%%%%%%%%%%%%%%%%%%%%
\begin{table*}[t]
\centering
\resizebox{\textwidth}{!}{
\begin{tabular}{cccccccc}
\toprule
 & ConvNet-IN &  & ConvNet-BN & ResNet18-IN & ResNet18-BN & VGG11-IN & VGG11-BN \\ \midrule
\multirow{3}{*}{\begin{tabular}[c]{@{}c@{}}Tiny\\ ImageNet\end{tabular}} & \multicolumn{1}{c}{\multirow{3}{*}{23.11 $\pm$ 1.83}} & \multicolumn{1}{c}{MTT~\cite{cazenavette2022dataset}} & 19.59 $\pm$ 0.27 & 10.02 $\pm$ 0.15 & 16.04 $\pm$ 0.56 & 11.23 $\pm$ 0.28 & 16.32 $\pm$ 0.47 \\
 & \multicolumn{1}{c}{} & \multicolumn{1}{c}{w. ELF} & 23.80 $\pm$ 0.30 & 19.73 $\pm$ 0.31 & 21.24 $\pm$ 0.49 & 19.56 $\pm$ 0.30 & 20.60 $\pm$ 0.66 \\ 
 & \multicolumn{1}{c}{} & \multicolumn{1}{c}{Gain} & $+$ 4.21 & $+$ 9.71 & $+$ 5.20 & $+$ 8.33 & $+$ 4.28 \\ \midrule
Full Dataset & 40.89 &  & 45.98 & 35.63 & 46.93 & 39.78 & 53.59 \\ \toprule
\multirow{3}{*}{ImageNet} & \multicolumn{1}{c}{\multirow{3}{*}{17.8 $\pm$ 1.3}} & \multicolumn{1}{c}{TESLA~\cite{cui2022scaling}} & 18.62 $\pm$ 0.10 & 8.83 $\pm$ 0.15 & 13.67 $\pm$ 0.05 & 14.76 $\pm$ 0.08 & 17.57 $\pm$ 0.14 \\
 & \multicolumn{1}{c}{} & \multicolumn{1}{c}{w. ELF} & 19.01 $\pm$ 0.07 & 15.65 $\pm$ 0.16 & 17.32 $\pm$ 0.19 & 19.96 $\pm$ 0.02 & 19.56 $\pm$ 0.38 \\
 & \multicolumn{1}{c}{} & \multicolumn{1}{c}{Gain} & $+$ 0.39 & $+$ 6.82 & $+$ 3.65 & $+$ 5.20 & $+$ 1.99 \\ \midrule
Full Dataset & 38.74 &  & 41.46 & 37.83 & 49.02 & 48.11 & 54.44 \\ \bottomrule
\end{tabular}}

\caption{Cross-architecture performance comparison on Tiny ImageNet and ImageNet-1K with 10 IPC.}
\label{table:ELF_cross_2}
\vspace{-10pt}
\end{table*}
%%%%%%%%%%%%%%%%%%%%%%%%%%%%%%%%%%%%%%%%%%%%%%%%%%%%%%%%%%%

%%%%%%%%%%%%%%%%%%%%%%%%%%%%%%%%%%%%%%%%%%%%%%%%%%%%%%%
\begin{figure*}[htbp]
    \centering
	\begin{minipage}{0.45\linewidth}
     \vspace{-1.3cm}
	\begin{table}[H]
	\centering
	\small
    \begin{tabular}{ccc}
    \toprule
	IPC & 1 & 10 \\ \midrule
	Baseline (ConvNet) & 26.52 $\pm$ 0.34 & 43.69 $\pm$ 0.27 \\
	w. ResNet Feature & 22.75 $\pm$ 1.29 & 34.05 $\pm$ 0.10 \\
	w. ConvNet Feature & {\bf 30.76 $\pm$ 0.34} & {\bf 48.60 $\pm$ 0.23} \\ \midrule
	Baseline (ResNet) & 10.12 $\pm$ 0.68 & 28.03 $\pm$ 0.26 \\
	w. ResNet Feature & 17.87 $\pm$ 0.65 & 32.58 $\pm$ 0.23 \\
	w. ConvNet Feature & {\bf 22.98 $\pm$ 0.42} & {\bf 40.34 $\pm$ 0.23} \\ \bottomrule
	\end{tabular}
	\caption{Performances of using different features on different baseline networks. ConvNet denotes ConvNetW512-IN and ResNet denotes ResNet18-IN in here. CIFAR-100 1/10 IPC, using ZCA preprocessing during distillation. Our default setting is highlighted in gray.}
    \label{tab:ablation_KD}
    \end{table}
	\end{minipage}
	%\qquad
	\hfill
	\begin{minipage}{0.45\linewidth}
		\begin{figure}[H]
      	\centering
      	\includegraphics[width=1\linewidth]{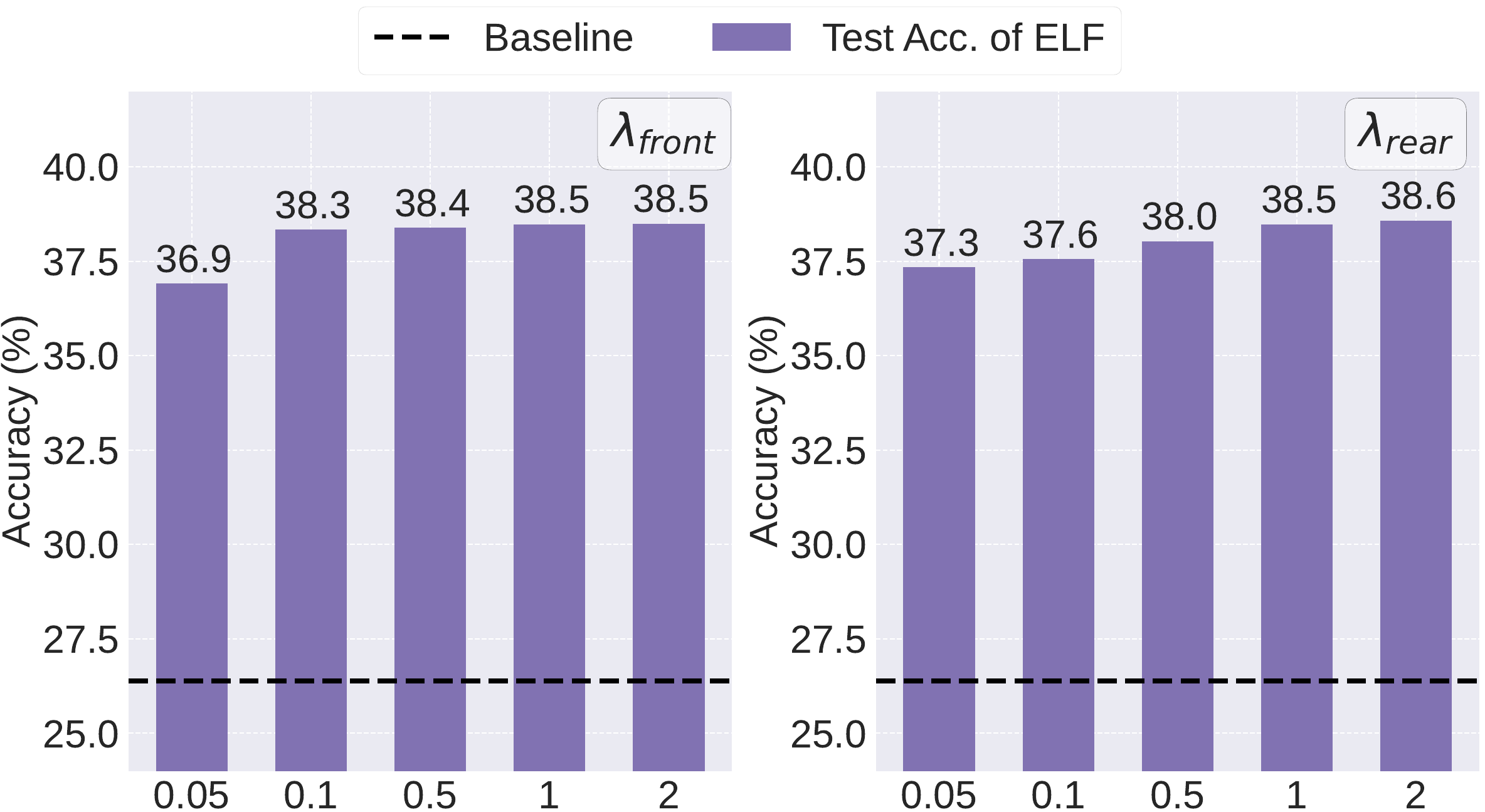}
      	\caption{Impacts of different $\lambda_{front}$ and $\lambda_{rear}$ on the test accuracy. CIFAR-100 10 IPC evaluated on ResNet18-IN.}
      	\label{fig:ablation_Loss_hyperparameter}
    	\end{figure}
    	\vspace{10mm}
	\end{minipage}
	\vspace{-1.5cm}
\end{figure*}
%%%%%%%%%%%%%%%%%%%%%%%%%%%%%%%%%%%%%%%%%%%%%%%%%%%%%%%

\section{Experimentation}\label{experiment}

\subsection{Experimental Details}\label{sec:4-1}
We evaluate our method on four standard image classiﬁcation benchmarks including CIFAR10/100~\cite{krizhevsky2009learning}, Tiny ImageNet~\cite{le2015tiny}, and ImageNet-1K~\cite{russakovsky2015imagenet}.
CIFAR-10/100 comprises 50,000 training images from 10/100 classes, with a resolution of 32$\times$32.
Tiny ImageNet dataset is composed of 100,000 training examples and 10,000 test examples with a higher resolution of 64$\times$64. They are from 200 classes.
ImageNet-1K is a widely-used large-scale dataset containing 1,000 classes and 1,281,167 training images.
%。
Following~\cite{zhou2022dataset,cui2022scaling}, the resolution in ImageNet-1K is resized to 64$\times$64.

We verify the efficacy of ELF in boosting the cross-architecture performance of representative DD methods, including DM~\cite{zhao2023dataset}, DSA~\cite{zhao2021dataset}, MTT~\cite{cazenavette2022dataset}, and TESLA~\cite{cui2022scaling}.
During the distillation phase, we maintain the same experimental settings as the baseline methods and use 3-/4-layer ConvNets as the distillation model for 32$\times$32/64$\times$64 resolution datasets respectively.
More details for the implementation of based methods are provided in the supplementary materials.
As to the cross-architecture generalization, the split of the ResNet-18 is located at the conv5\_2 for CIFAR-10/100 and conv4\_4 for Tiny ImageNet and ImageNet-1K.
Meanwhile, we procure the bias-free intermediate features from ConvNet with widths of 256 and 512 to respectively assess datasets with 32$\times$32 and 64$\times$64 resolution.
For VGG11, we conduct the splitting at its fifth/sixth layer and extract the bias-free intermediate features from ConvNet with a width of 512 for 32$\times$32/64$\times$64 resolution datasets, respectively.
Besides, we choose the cross-entropy loss for the distance function in Eq.\,(\ref{eq:front_loss}).
$\lambda_{front}$ and $\lambda_{rear}$ in Eq.\,(\ref{eq:loss}) are uniformly set to 1.
We run each experiment five times and report the mean and standard deviation of the results.
All experiments are implemented using PyTorch~\cite{paszke2019pytorch} and executed on four NVIDIA 3090 GPUs.

%%%%%%%%%%%%%%%%%%%%%%%%%%%%%%%%%%%%%%%%%%%%%%%%%%%%%%%%%%%
\begin{table*}[!t]
\centering
\begin{tabular}{ccccccc}
\toprule
&$\mathcal{L}_{task}$ & $\mathcal{L}_{front}$ & $\mathcal{L}_{rear}$ & 1 & 10 & 50 \\ \midrule
Baseline & \cmark & - & - & 12.43 $\pm$ 0.99 & 26.39 $\pm$ 0.66 & 39.67 $\pm$ 0.61 \\
(a)& \cmark & - & \cmark & 14.05 $\pm$ 0.37 & 27.77 $\pm$ 0.96 & 41.98 $\pm$ 0.19 \\
(b)& \cmark & \cmark & - & 21.47 $\pm$ 0.28 & 37.37 $\pm$ 0.15 & 47.80 $\pm$ 0.19 \\
(c)& - & \cmark & \cmark & 16.05 $\pm$ 0.59 & 17.32 $\pm$ 2.54 & 23.24 $\pm$ 3.07 \\
ELF&\cmark & \cmark & \cmark & \textbf{21.73 $\pm$ 0.30} & \textbf{38.48 $\pm$ 0.14} & \textbf{48.45 $\pm$ 0.22} \\ \bottomrule
\end{tabular}
\caption{Results of ablation study on loss terms in ELF: $\mathcal{L}_{task}$, $\mathcal{L}_{front}$, and $\mathcal{L}_{rear}$. (CIFAR-100, ResNet18-IN)}
\label{tab:ablation_Loss_combination}
\vspace{-0.2cm}
\end{table*}
%%%%%%%%%%%%%%%%%%%%%%%%%%%%%%%%%%%%%%%%%%%%%%%%%%%%%%%%%%%

%%%%%%%%%%%%%%%%%%%%%%%%%%%%%%%%%%%%%%%%%%%%%%%%%%%%%%%%%%%
\begin{figure*}[!t]
\begin{center}

\includegraphics[width = 0.9\linewidth]{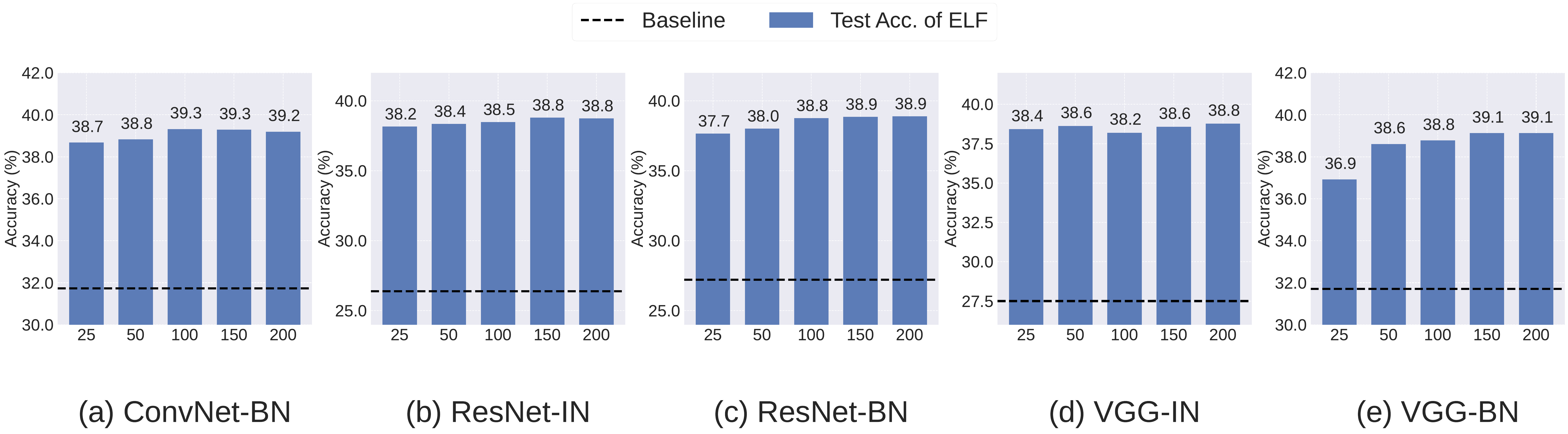}
\end{center}
\vspace{-0.2cm}
\caption{Ablation studies on feature epoch in ELF. The horizontal coordinate denotes the number of epochs learned by the distillation model that generated the required features.}
\label{fig:ablation_epoch}
\vspace{-0.4cm}
\end{figure*}
%%%%%%%%%%%%%%%%%%%%%%%%%%%%%%%%%%%%%%%%%%%%%%%%%%%%%%%%%%%

\subsection{Quantitative Comparison}\label{sec:4-2}

\textbf{CIFAR-10/100}. Table\,\ref{table:ELF_cross_0} and Table\,\ref{table:ELF_cross_1} respectively display the cross-architecture performance of DD methods with or without the aid of ELF on CIFAR-10 and CIFAR-100 datasets.
As it can be inferred, all DD methods yield poor cross-architecture performance due to the presence of inductive bias.
Excitingly, by utilizing features from intermediate layers of the distillation model, ELF consistently improves the cross-architecture performance of DD methods by a large margin.
For instance, ELF leads to a substantial 12.19\% enhancement in the accuracy of MTT when evaluating VGG11-IN with 10 images distilled from CIFAR-10. 
Moreover, the implementation of ELF greatly enhances the cross-architecture performance of DM, exhibiting results commensurate with that of ConvNet-IN despite its lack of inductive bias.
These outcomes effectively corroborate our standpoint on boosting the cross-architecture generalization of DD method from the perspective of eliminating inductive bias.

\textbf{Tiny-ImageNet/ImageNet-1K}. We further investigate the efficacy of ELF on datasets with larger scale,~\emph{i.e.}, Tiny-ImageNet and ImageNet-1K.
The results are listed in Table\,~\ref{table:ELF_cross_2}.
It is apparent that ELF continues to exhibit remarkable superiority when confronted with larger datasets.
With respect to Tiny-ImageNet, ELF consistently improves the performance of MTT with accuracies ranging from 4.21\% to 9.71\% for the cross-architecture evaluation of multiple networks.
The same conclusion can also be drawn when it comes to ImageNet-1K using TESLA~\cite{cui2022scaling} as the baseline method.
TESLA assigns soft labels on DD to distill generalizable information to synthetic datasets.
The results show that ELF can be effectively combined with other techniques to jointly enhance cross-architecture performance.
In summary, ELF demonstrates its great capability to boost the cross-architecture performance of DD on the large scale datasets.

\subsection{Performance Analysis}\label{sec:4-3}

\textbf{Distillation feature}.
To demonstrate the effectiveness of the distillation feature in  mitigating the inductive bias, we replace the features used in Eq.\,(\ref{eq:loss}) with features extracted from trained evaluation model on the full dataset.
In detail, we set ConvNetW512 as the distillation model and extract features from either ResNet18 or ConvNetW512 trained on the full dataset to evaluate the performances of ConvNetW512 and ResNet18 respectively.
The baseline method for performance analysis is MTT~\cite{cazenavette2022dataset} if not specified.
Surprisingly, Table\,\ref{tab:ablation_KD} shows that utilizing the distillation model's features yields significantly superior results compared with utilizing the evaluation model's features.
This strongly proves the importance of eliminating the inductive bias with bias-free features in the cross-architecture evaluation of DD.

\textbf{Loss terms}.
Next, we perform ablation studies to investigate the efficacy of each loss term in Eq.\,(\ref{eq:loss}).
Table\,\ref{tab:ablation_Loss_combination} shows that all loss terms of ELF play a unique role in mitigating the influence of the inductive bias.
Further, Fig.\,\ref{fig:ablation_Loss_hyperparameter} investigates the influence of hyperparameters $\lambda_{front}$ and $\lambda_{rear}$ corresponding to $\mathcal{L}_{front}$ and $\mathcal{L}_{rear}$ in Eq.\,\ref{eq:loss}.
As can be seen, the performance of ELF is also robust to the hyper-parameter setting of loss.

\textbf{Features from different epoch}.
At last, we investigate the number of epochs necessary for distillation model features.
The results are evaluated on the CIFAR-100 dataset 10 IPC with different evaluation models. 
As depicted in Fig.\,\ref{fig:ablation_epoch}, ELF's performance is relatively insensitive to the choice of epoch for the extracted features.
This allows us to obtain the required features with significantly fewer resources compared to the process of DD itself.

\section{Limitation}
Although ELF has been shown to effectively enhance the cross-architecture performance of DD, it still presents some unexplored limitations.
Firstly, the utilization of features from the distillation model is subject to some degree of shape restriction, which limits its applicability in some extreme scenarios.
For instance, in cases where the evaluation models require very large feature maps, such as 32$\times$32, obtaining features of the necessary size may not be feasible from the distillation model trained on the small-scale CIFAR-10/100 datasets.
Furthermore, ELF introduces some additional storage burdens for the distillation features. 
While this does not align with our current goal of enhancing cross-architecture performance, it would be worthwhile to explore ways to reduce this extra storage burden in future work.

\section{Conclusion}\label{Conclusions}
This paper proposes ELF, a novel method for enhancing the cross-architecture generalization of dataset distillation (DD).
ELF utilizes the features extracted from the distillation model to steer the training of the evaluation model on the synthetic dataset generated by DD.
Extensive experiments on multiple datasets demonstrate that ELF can effectively enhance the cross-architecture performance of existing DD methods.
Unlike previous works, this paper provides a specific focus on enhancing the cross-architecture performance of DD methods, while also undertaking a comprehensive analysis and experimentation on their cross-architecture generalization.
We believe that this work could inspire future research enthusiasm on more exploration toward the practical applications of DD.

\bibliography{aaai24}

\clearpage
\begin{appendix}

\section{Experimental Details}
We provide more details regarding our implementation of DD methods.
We acquire the synthetic dataset and distillation feature by utilizing the officially released code of each respective method~\footnote{https://github.com/VICO-UoE/DatasetCondensation}
\footnote{https://github.com/GeorgeCazenavette/mtt-distillation}
\footnote{https://openreview.net/forum?id=dN70O8pmW8}.
For a fair comparison, we evaluate the performance of all synthetic datasets using 1000 training epochs with a learning rate of 0.01.
In what follows, we present more experimental details for each table in the main manuscript.

\textbf{Table 1-4}. We generate the synthetic dataset for each method using their default settings. 
For MTT~\cite{cazenavette2022dataset}, we do not employ ZCA whitening in order to maintain consistency of different DD methods~\cite{zhao2021dataset,zhao2023dataset}. 
Moreover, we notice that using ZCA whitening even reduces the cross-architecture performance of MTT in small IPC.

\textbf{Table 5-7}. The cross-architecture performance comparison between ELF and the corresponding baseline method is based on evaluations using the same synthetic dataset.
Furthermore, Figure\,\ref{fig:ablation_epoch} highlights that the performance of ELF remains relatively stable across epochs for feature extraction. 
Hence we only select the coarse-grained epoch of features in different evaluation settings.
The epochs used to obtain the features are listed in Table\,\ref{tab:settings_of_epoch}.
For DM and DSA in Table\,\ref{table:ELF_cross_0}, we set the learning rate to 0.005 instead of 0.01 because of significant loss fluctuation at large learning rates in our experimental observation.

%%%%%%%%%%%%%%%%%%%%%%%%%%%%%%%%%%%%%%%%%%%%%%%%%%%%%%%%%%%%%%%%
\begin{table}[h]
    \centering
    \begin{tabular}{l|cccc}
    \toprule
     10 IPC & DM & DSA & MTT & TESLA \\ \midrule
    Table\,\ref{table:ELF_cross_0} (CIFAR-10) & 30 & 30 & 50 & - \\
    Table\,\ref{table:ELF_cross_1} (CIFAR-100)& 50 & 50 & 100 & - \\
    Table\,\ref{table:ELF_cross_2} (Tiny-ImageNet) & - & - & 100 & - \\
    Table\,\ref{table:ELF_cross_2} (ImageNet-1K) & - & - & - & 100 \\ \bottomrule
    \end{tabular}
    
    \caption{Epochs used to get the features from the distillation model in ELF.}
    \label{tab:settings_of_epoch}
    
\end{table}
%%%%%%%%%%%%%%%%%%%%%%%%%%%%%%%%%%%%%%%%%%%%%%%%%%%%%%%%%%%%%%%%

\textbf{Table 8}. In Table\,\ref{tab:ablation_KD} of the main paper, we employ ZCA whitening. 
We also present experimental results under the same settings but without ZCA whitening in Table\,\ref{tab:ablation_KD_no_zca}.
As can be seen, the trend remains consistent that the results obtained using the distillation model's feature exceed those obtained using the evaluation model's feature.
Overall, the significance of removing the inductive bias using bias-free features in the cross-architecture evaluation of DD is apparent.
% 

%%%%%%%%%%%%%%%%%%%%%%%%%%%%%%%%%%%%%%%%%%%%%%%%%%%%%%%%%%%%%%%%
\begin{table}[h]
    \centering
    \begin{tabular}{ccc}
	\toprule
	IPC & 1 & 10 \\ \midrule
	Baseline (ConvNet) & 18.13$\pm$0.47 & 38.49 $\pm$ 0.35 \\
	w. ResNet Feature & 15.36 $\pm$ 0.63 & 32.81 $\pm$ 0.31 \\
	  w. ConvNet Feature & {\bf 26.84 $\pm$ 0.57 } & {\bf 46.29 $\pm$ 0.34} \\ \midrule
	Baseline (ResNet) & 8.49 $\pm$ 0.60 & 24.40 $\pm$ 0.31 \\
	w. ResNet Feature & 15.04 $\pm$ 0.18 & 30.60 $\pm$ 0.21 \\
	w. ConvNet Feature & {\bf 18.87 $\pm$ 0.16 } & {\bf 36.31      $\pm$ 0.25} \\ \bottomrule
    \end{tabular}

    \caption{Performances of using different features on different baseline networks. CIFAR-100 1/10 IPC, not using ZCA preprocessing during distillation.}
\label{tab:ablation_KD_no_zca}
\end{table}
%%%%%%%%%%%%%%%%%%%%%%%%%%%%%%%%%%%%%%%%%%%%%%%%%%%%%%%%%%%%%%%%

\section{More Performance Analysis}

%%%%%%%%%%%%%%%%%%%%%%%%%%%%%%%%%%%%%%%%%%%%%%%%%%%%%%%%%%%
\begin{table*}[!b]
\centering
\begin{tabular}{clccccc}
\toprule
&  & ConvNet-BN & ResNet-IN & ResNet-BN & VGG-IN & VGG-BN \\ \midrule
\multirow{4}{*}{DSA} & Baseline & 27.56 $\pm$ 0.18 & 21.96 $\pm$ 0.51 & 20.45 $\pm$ 0.53 & 22.00 $\pm$ 0.34 & 25.73 $\pm$ 0.41 \\
& MAE & 34.32 $\pm$ 0.24 & 22.47 $\pm$ 0.75 & 25.57 $\pm$ 0.60 & 24.73 $\pm$ 0.38 & 28.37$\pm$0.37 \\
& MSE & 35.42 $\pm$ 0.22 & 21.89 $\pm$ 1.21 & 25.62 $\pm$ 0.57 & 25.03 $\pm$ 0.26 & 28.65 $\pm$ 0.37 \\
& Cos & 34.57 $\pm$ 0.23 & 24.07 $\pm$ 0.11 & 27.31 $\pm$ 0.68 & 25.21 $\pm$ 0.32 & \textbf{30.05 $\pm$ 0.37} \\
& CE & \textbf{36.02 $\pm$ 0.35} & \textbf{27.54 $\pm$ 0.19} & \textbf{30.26 $\pm$ 0.65} & \textbf{28.74 $\pm$ 0.11} & 28.54 $\pm$ 1.23 \\ \midrule
\multirow{4}{*}{MTT} & Baseline & 31.73 $\pm$ 0.15 & 26.39 $\pm$ 0.66 & 27.21 $\pm$ 0.53 & 27.50 $\pm$ 0.26 & 31.71 $\pm$ 0.58 \\
& MAE & 39.45 $\pm$ 0.30 & 28.01 $\pm$ 0.76 & 33.10 $\pm$ 0.55 & 33.04 $\pm$ 0.41 & 33.80 $\pm$ 0.86 \\
& MSE & \textbf{40.07 $\pm$ 0.49} & 26.98 $\pm$ 2.00 & 33.43 $\pm$ 0.56 & 33.41 $\pm$ 0.53 & 34.59 $\pm$ 0.75 \\
& Cos & 39.63 $\pm$ 0.34 & 31.10 $\pm$ 0.53 & 35.53 $\pm$ 0.47 & 33.89 $\pm$ 0.32 & 36.49 $\pm$ 0.39 \\
& CE & 39.32 $\pm$ 0.24 & \textbf{38.48 $\pm$ 0.14} & \textbf{38.76 $\pm$ 0.80} & \textbf{38.20 $\pm$ 0.49} & \textbf{38.78 $\pm$ 0.84} \\ \bottomrule
\end{tabular}
\caption{Results of ablation study on distance function in Eq.\,(\ref{eq:front_loss}). CIFAR-100 10 IPC in DSA~\cite{zhao2021dataset} and MTT~\cite{cazenavette2022dataset}. \emph{MAE} denotes mean-absolute-error distance function, \emph{MSE} denotes mean-square-error distance function, \emph{Cos} denotes cosine-similarity distance function, and \emph{CE} denotes cross-entropy distance function. The Cross-architecture performance using CE loss is much better than using others.}
\label{tab:ablation_loss_function_selection}
\end{table*}
%%%%%%%%%%%%%%%%%%%%%%%%%%%%%%%%%%%%%%%%%%%%%%%%%%%%%%%%%%%

\textbf{Loss function.}
We further study the effect of the selection of distance function $D(\cdot,\cdot)$ in Eq.\,\ref{eq:front_loss}.
Table\,\ref{tab:ablation_loss_function_selection} shows that the cross-architecture performance when using the cross-entropy (CE) loss as $D(\cdot,\cdot)$ is significantly better than when using other loss functions, including mean-absolute-error (MAE), mean-square-error (MSE), and cosine-similarity (Cos).

\end{appendix}

\end{document}